%% file: main.tex
\documentclass[10pt,twocolumn,letterpaper]{article}

\usepackage{cvpr}              

\input{preamble}

\definecolor{cvprblue}{rgb}{0.21,0.49,0.74}
\usepackage[pagebackref,breaklinks,colorlinks,allcolors=cvprblue]{hyperref}

\def\paperID{*****} 
\def\confName{CVPR}
\def\confYear{2025}

\title{YETI (YET to Intervene) Proactive Interventions by Multimodal AI Agents in Augmented Reality Tasks}

\author{Saptarashmi Bandyopadhyay\\
University of Maryland, College Park\\
College Park, MD, USA\\
{\tt\small saptab1@umd.edu}
\and
Vikas Bahirwani\\
Google\\
Mountain View, CA, USA\\
{\tt\small vrb@google.com}
\and
Lavisha Aggarwal\\
Google\\
Seattle, WA, USA\\
{\tt\small lavishaggarwal@google.com}
\and
Bhanu Guda\\
Google\\
Mountain View, CA, USA\\
{\tt\small bhanuguda@google.com}
\and
Lin Li\\
Google\\
Mountain View, CA, USA\\
{\tt\small linspeaking@google.com}
\and
Andrea Colaco\\
Google\\
Mountain View, CA, USA\\
{\tt\small andreacolaco@google.com}}

\begin{document}
\maketitle
\input{sec/0_abstract}   
\input{sec/1_introduction}

\input{sec/2_related_works}
\input{sec/3_methodology}

\input{sec/4_experiments}
\input{sec/5_conclusion}

{
    \small
    \bibliographystyle{ieeenat_fullname}
    \bibliography{main}
}

\input{sec/act_supplementary}



\end{document}

%% file: preamble.tex
\usepackage{algorithm}  
\usepackage{algpseudocode}  
\usepackage{amsmath}  
\usepackage{amsthm}  
\usepackage{mathtools}  

\usepackage{pifont}

\usepackage{multirow}    

\newcommand\blfootnote[1]{
    \begingroup
    \renewcommand\thefootnote{}\footnote{#1}
    \addtocounter{footnote}{-1}
    \endgroup
}

%% file: sec/0_abstract.tex
\begin{abstract}
Multimodal AI Agents are AI models that have the capability of interactively and cooperatively assisting human users to solve day-to-day tasks. Augmented Reality (AR) head worn devices can uniquely improve the user experience of solving procedural day-to-day tasks by providing egocentric multimodal (audio and video) observational capabilities to AI Agents. Such AR capabilities can help the AI Agents see and listen to actions that users take which can relate to multimodal capabilities of human users. Existing AI Agents, either Large Language Models (LLMs) or Multimodal Vision-Language Models (VLMs) are reactive in nature, which means that models cannot take an action without reading or listening to the human user's prompts. Proactivity of AI Agents on the other hand can help the human user detect and correct any mistakes in agent observed tasks, encourage users when they do tasks correctly or simply engage in conversation with the user - akin to a human teaching or assisting a user. Our proposed YET to Intervene (YETI) multimodal agent focuses on the research question of identifying circumstances that may require the agent to intervene proactively. This allows the agent to understand when it can intervene in a conversation with human users that can help the user correct mistakes on tasks, like cooking, using Augmented Reality. Our YETI Agent learns scene understanding signals based on interpretable notions of Structural Similarity (SSIM) on consecutive video frames. We also define the alignment signal which the AI Agent can learn to identify if the video frames corresponding to the user's actions on the task are consistent with expected actions. These signals are used by our AI Agent to determine when it should proactively intervene. We compare our results on the instances of proactive intervention in the HoloAssist multimodal benchmark for an expert agent guiding a user to complete procedural tasks.\blfootnote{Preprint.}
\end{abstract}

%% file: sec/1_introduction.tex
\section{Introduction}
\label{sec:introduction}
Recent advances in artificial intelligence have led to the widespread adoption of AI assistants across various platforms and modalities. While these systems, such as Siri for voice interaction and Gemini~\cite{team2023gemini} for text-based communication, have demonstrated significant utility in task automation, they remain constrained by their single-modality architectures. This limitation presents a critical gap in human-AI interaction, particularly in scenarios requiring real-time, context-aware assistance.

Multimodal Vision-Language Models (VLMs) have emerged as a promising solution to bridge this modality gap, offering multimodal understanding that more closely aligns with human perception. However, current VLM-based assistive systems predominantly operate in a reactive paradigm, responding only to explicit user queries. This reactive nature significantly limits their effectiveness in two critical scenarios: (1) novice learning environments, where users lack the domain knowledge to formulate appropriate queries, and (2) safety-critical operations, where immediate intervention may be necessary before user recognition of potential hazards.

\begin{figure*}[!ht]
    \centering
    \includegraphics[width=\textwidth]{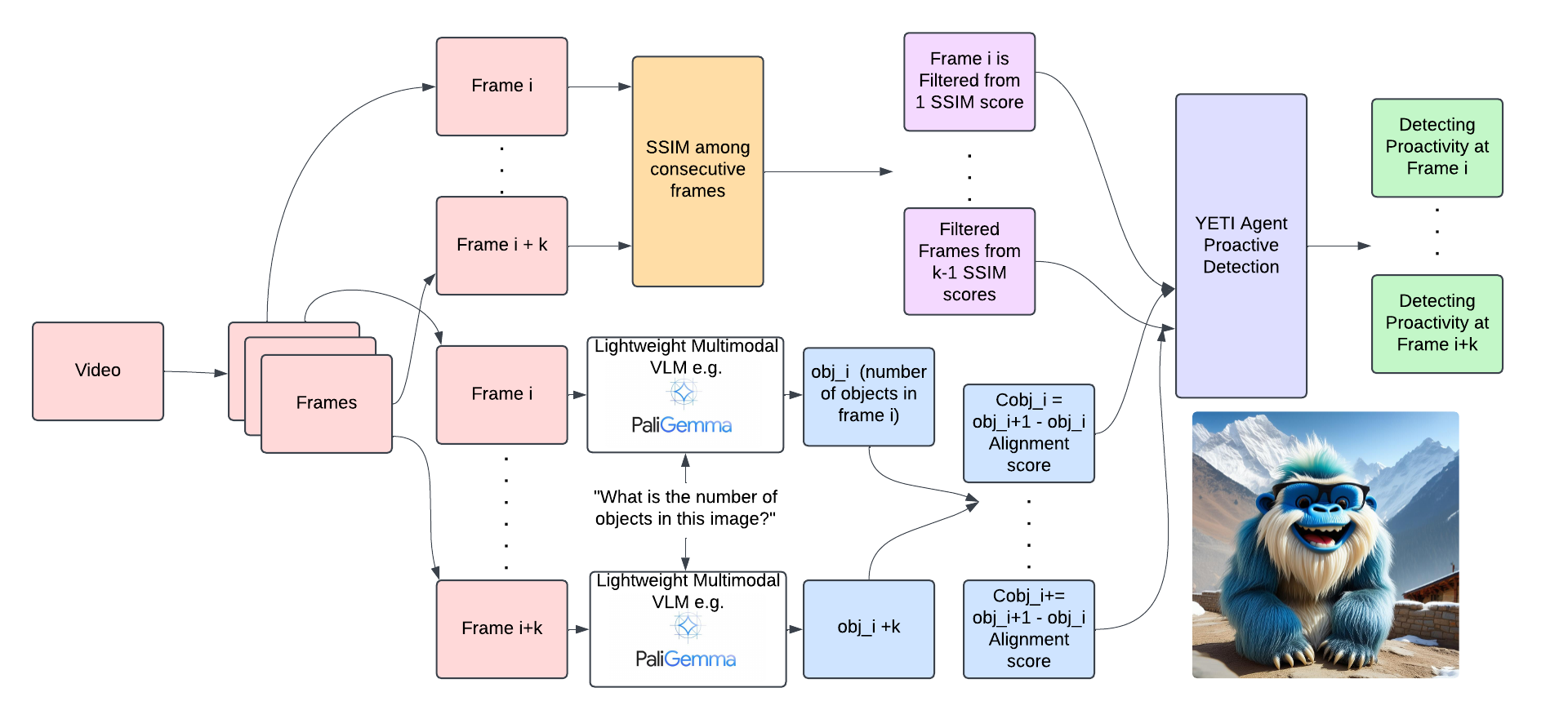}
        \caption{Overview of the YETI framework detecting the frames of proactive interaction or intervention by a Multimodal AI Agent. Our YETI Agent system generates lightweight features on-the-fly, enabling rapid  decision-making for timely user assistance.}
    \label{fig:system_overview}
\end{figure*}

To address these limitations, we propose \textbf{YET} to \textbf{I}ntervene (YETI), a novel framework (seen in Figure \ref{fig:system_overview}) for proactive AI intervention in augmented reality (AR) environments. Our approach leverages lightweight, real-time algorithmic signals to enable proactive assistance through AR interfaces such as smart glasses~\cite{waisberg2024meta}. This system bridges the gap between cloud-based AI capabilities and real-world applications by enabling direct visual observation of user activities.

Our work builds upon recent developments in proactive AI assistance, particularly the HoloAssist dataset~\cite{wang2023holoassist}, which demonstrates the potential for real-time AI intervention in complex spatio-temporal tasks. While HoloAssist provides valuable insights into human-AI collaboration scenarios, such as computer assembly and coffee preparation, existing implementations face significant computational challenges.

Current state-of-the-art approaches for proactive intervention require extensive computational resources and multi-modal sensor data, including RGB streams, hand and head pose estimation, sensor readings like IMU (Inertial Measurement Unit), and depth information. The complexity of acquiring and processing this data in real-time presents a significant barrier to practical deployment. In contrast, YETI employs efficient algorithmic signals that can be computed on-the-fly, dramatically reducing the computational overhead while maintaining high intervention accuracy.

\begin{table}[h]
\centering
\begin{tabular}{p{0.3\linewidth}ccc}
\toprule
Features & Size (MB) & {$\times$ SSIM} & {$\times$ CObj}\\
\midrule
Depth Estimation & 137,408 & 6,543 & 6,870 \\
Eye Gaze (E)& 617 & 29 & 31\\
Hand Pose (H) & 53,749 & 2,660 & 2,688\\
Head Pose & 1,141 & 54 & 57\\
IMU (I) & 1,132 & 54 & 57 \\
\midrule
\textbf{SSIM} (Ours) & \textbf{21} \\
\textbf{Alignment Cobj} (Ours) & \textbf{20} \\
\bottomrule
\end{tabular}
\caption{HoloAssist Feature sizes scaled with our Features}
\label{table:data_sizes}
\end{table}




The YETI framework has comparable precision performance with different HoloAssist benchmark baseline models and shows better performance in some settings, specially higher recall and F-measure, all while using light-weight features that take 6500 times less memory, as seen in Table \ref{table:data_sizes}. This dramatic reduction in computational requirements enables real-time operation on resource-constrained AR devices, making proactive AI assistance practical for everyday use cases. Our framework thus represents a significant step toward deploying intelligent assistive systems in real-world applications, particularly in scenarios requiring immediate, context-aware intervention.

%% file: sec/2_related_works.tex
\section{Related Works}
\label{sec:related_works}

\subsection{Egocentric Interaction Datasets}
Recent advances in egocentric vision have produced several datasets that capture human interactions and activities. HoloAssist~\cite{wang2023holoassist} presents a large-scale egocentric dataset focusing on collaborative physical manipulation tasks between two people, providing detailed action and conversational annotations. This dataset offers valuable insights into how human assistants proactively and reactively intervene, correct mistakes, and ground their instructions in the environment.

Parse-Ego4D~\cite{abreu2024parseego4dpersonalactionrecommendation} introduces a benchmark for evaluating AI agents' capability to make unsolicited action suggestions based on user intent signals. We argue that as this benchmark evaluates the AI agents' response to user queries, it does not truly measure proactive behavior.

While Ego-Exo4D~\cite{grauman2024egoexo4dunderstandingskilledhuman} provides a comprehensive multimodal, multiview dataset capturing both egocentric and exocentric perspectives in expert-learner scenarios, it primarily focuses on skilled single-person activities without addressing proactive communication. Similarly, existing datasets like Ego4D~\cite{Grauman_2022_CVPR} and EPIC-Kitchens \cite{Damen_2018_ECCV}, while rich in activity and object annotations, lack direct mappings to actionable recommendations.

\subsection{Proactive AI Agents and Communication}
Proactive communication in AI agents encompasses several key aspects~\cite{10.1145/3626772.3657843}: Intelligence (the ability to anticipate task developments), Adaptivity (dynamic adjustment of timing and interventions), and Civility (respect for user boundaries and ethical standards). Emerging research has demonstrated the value of proactive AI agents across various domains, including personal assistance, predictive maintenance, healthcare monitoring, and voice assistance~\cite{BERUBE2024100411,butala-etal-2024-promise,10.1145/3539618.3594250}.

\subsection{Language Models for Proactive Assistance}
Recent developments have shown promising results in using Multimodal Vision Language Models (VLMs) and LLMs for proactive assistance. ProAgent~\cite{Zhang_Yang_Hu_Wang_Li_Sun_Zhang_Zhang_Liu_Zhu_Chang_Zhang_Yin_Liang_Yang_2024} introduces a framework that leverages LLMs to create agents capable of dynamically adapting their behavior and inferring teammate intentions. The effectiveness of these models has been further demonstrated through fine-tuning on ProactiveBench~\cite{lu2024proactiveagentshiftingllm}, which significantly enhances the proactive capabilities of LLM agents. In the context of assistive technology, Smart Help~\cite{Cao_2024_CVPR} demonstrates how proactive and adaptive support can be provided to users with diverse disabilities and dynamic goals across various tasks and environments.

Open-source VLMs are very popular as a starting point for AI assistants, especially Google's PaliGemma~\cite{beyer2024paligemmaversatile3bvlm} Open-source VLM. Open-source VLMs do not have proactive interaction capabilities which is what we want to support in our research. PaliGemma generates a quick and accurate estimate of the number of objects in a given scene. PaliGemma was trained on a wide variety of datasets, including the TallyQA dataset~\cite{acharya2019tallyqa}, which is useful for taking a response to a question that asks for the number of objects in a given image.

%% file: sec/3_methodology.tex
\section{Methodology}
\label{sec:method}

\subsection{Proactive Augmented Reality Interaction Data of Cooperative Agents}
\label{subsec:proactive_ai_agent_interaction}

The HoloAssist dataset \citep{wang2023holoassist} provides multimodal ego-centric vision-language benchmarks focusing on Augmented Reality (AR)-based human-AI collaboration. AR devices are used to capture the Expert-User collaborative dynamics, recording the visual observations of an User Agent (human) collaborating with an Expert Agent (Instructor), which can be an AI Agent, on physical reasoning tasks, while documenting the dialogue between the two agents. The dataset comprises 482 unique Expert-User interaction sequences with videos and dialogues of the agents', spanning 20 diverse task domains, including but not limited to:
\begin{itemize}
    \item Cooking procedures like making coffee
    \item Fixing items like motorcycles
    \item Assembling/Disassembling furniture
    \item Assembling Devices like Computers, Scanners, GPUs
    \item Maintaining Electrical systems like circuit breakers
    \item Configuring Devices like printers, cameras, switches
\end{itemize}

The User Agents wear the AR devices to record first-person perspective videos while executing procedural tasks. The AR devices simultaneously capture the Expert Agent's observations and guidance to the User Agents. The dataset's annotation schema encompasses a variety of interaction types. Some examples of interactions done by the Expert Agent include:



\begin{enumerate}
    \item Proactive Interactions
    \begin{itemize}
        \item High-level instructional guidance
        \item Follow-up instructions without any user query
        \item Interventional feedback
        \item Error correction mechanisms
    \end{itemize}
    \item Reactive interactions:
    \begin{itemize}
        \item Expert clarifications to user queries
        \item User-initiated dialogues
    \end{itemize}
\end{enumerate}

The corpus encompasses 45.5 hours of video recordings generated by 350 distinct expert-user pairs, providing a rich foundation to study the spatio-temporal dynamics of when and how Proactive Multimodal AI Agents should engage in AR-assisted collaborative scenarios. This temporal aspect, when an AI Agent is yet to intervene, but should intervene, to improve collaborative task execution or correct mistakes is crucial for developing AI agents which can guide humans in human-AI collaborative tasks.

\subsection{Multimodal VLM Generation}
\label{subsec:vlm_generation}

We count objects in video sequences by leveraging the recent advancements in Multimodal Visual Language Models (VLMs). Our method processes videos by extracting frames at a rate of 1 frame per second (FPS), enabling efficient temporal analysis while maintaining sufficient granularity for accurate object counting.

\paragraph{Frame Extraction and Processing}
Given an input video $V$ of duration $T$ seconds, we extract a sequence of frames $\{f_1, f_2, ..., f_T\}$ at 1 FPS. This sampling rate balances computational efficiency with temporal resolution, ensuring that significant object state changes are captured while minimizing redundant processing.

\begin{figure}[!h]
    \centering
    \hspace*{-1.5cm}
    \includegraphics[width=\linewidth]{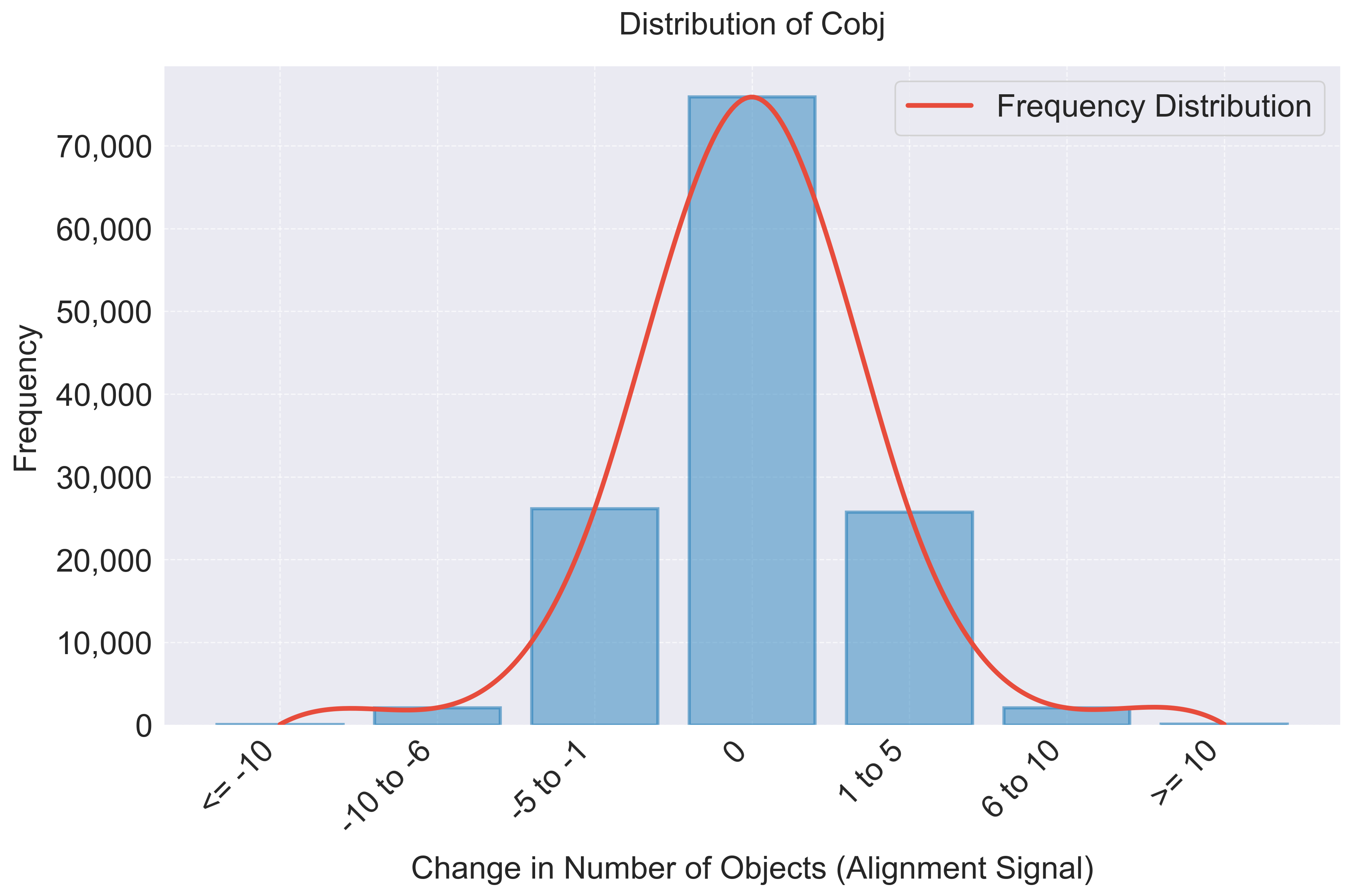}
    \caption{Distribution of Alignment Signal}
    \label{fig:histogram}
\end{figure}

\paragraph{Multimodal VLM Implementation}
We utilize PaliGemma-3b-mix-448, from the PaliGemma~\cite{beyer2024paligemmaversatile3bvlm} family of lightweight multimodal VLMs created by Google, to process each frame independently. PaliGemma's ability to quickly leverage its visual and textual understanding capabilities makes it suitable for an AR / VR setting where there may not be much device compute available and a fast response is needed.
For each frame $f_t$, we construct a prompt:
\begin{equation}
\label{eq:prompt}
    P_{obj} = \text{``The number of objects in this image is "}
\end{equation}
This prompt elicits a numerical response from the model, avoiding unwanted conversational output. We chose this prompt after thorough experimentation with PaliGemma's object detection capabilities. Some of the other prompts we explored as well as their output can be found in the supplementary material. The model processes each frame $f_t$ with prompt $P_{obj}$ to generate a count estimate:
\begin{equation}
\label{eq:generation}
    C_t = \text{PaliGemma}(f_t, P_{obj})
\end{equation}
where $C_t$ represents the predicted object count at time $t$. The change in object count between frames is heavily skewed towards zero as seen in Figure \ref{fig:histogram}.

\paragraph{Implementation Details}
The PaliGemma model was used with its default configuration, maintaining the $448\times448$ pixel input resolution. The model's responses were post-processed to extract numerical values.

\subsection{Alignment with Changing Object Count}
\label{subsec:alignment_cobj}

\begin{figure}[!ht]
    \centering
        \subfloat[Frame 36]{\includegraphics[width=0.22\textwidth]{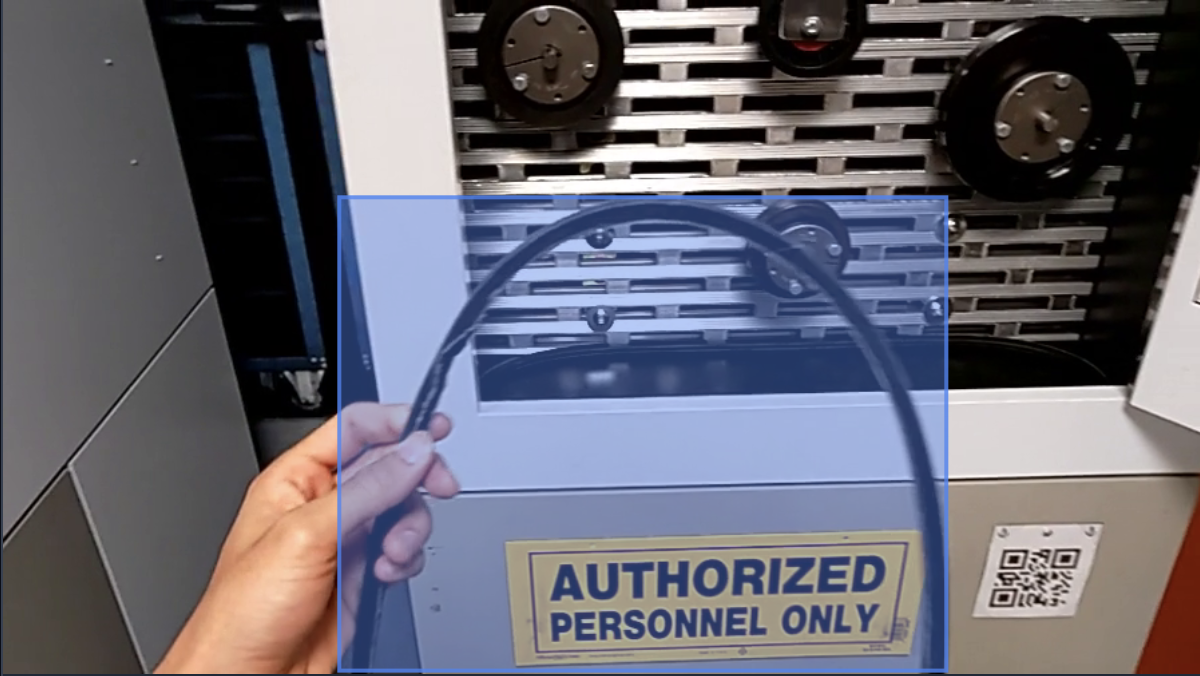}}
    \hfill
    \subfloat[Frame 37]{\includegraphics[width=0.22\textwidth]{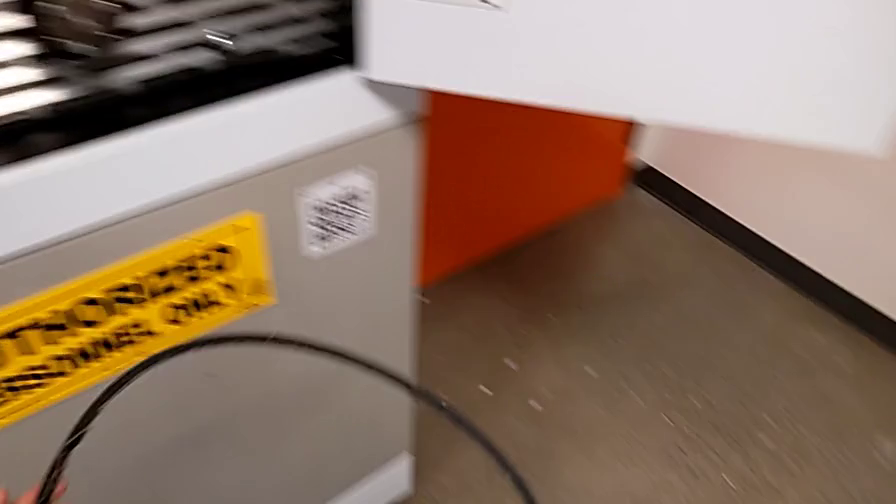}} \\
    \subfloat[Alignment Signal over entire video.]{\includegraphics[width=0.22\textwidth]{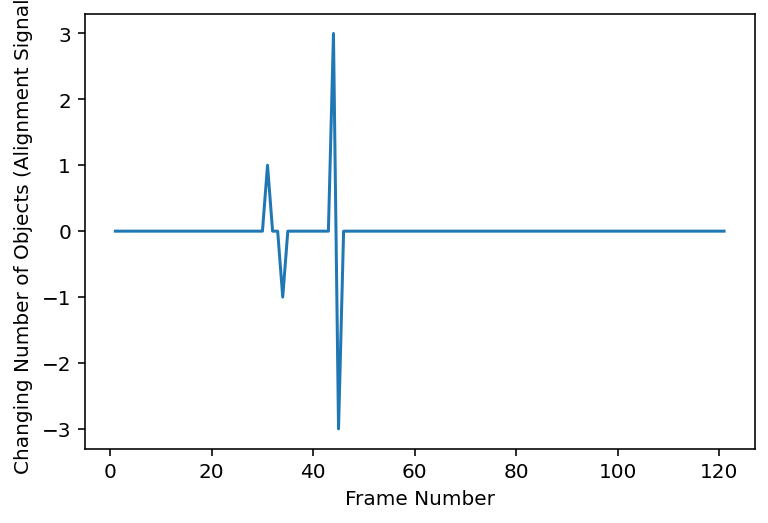}}
    \caption{Plot of Alignment Signal measuring changing object count along spatio-temporally changing image frames in a video for a procedural task on how to change a mechanical belt. The Expert Agent autonomously intervenes in the 37th second at Frame 37 based on the alignment signal with Frame 36}
    \label{fig:alignment_cobj_eg}
\end{figure}

The alignment signal taking the form of a change in object count for the scene observed by the AI Assistant is motivated by an intuition of how humans operate when listening to instructions. If a user agent is being guided on how to assemble a computer, they will not be moving objects around while they process the instructions. Rather, they will be listening so they know what to do next. Building upon this understanding, we can estimate when a proactive AI assistant should intervene by simply monitoring the change in object count from second to second. Figure \ref{fig:alignment_cobj_eg} shows an example of frames where the object count changes.

\subsection{Scene Understanding}
\label{subsec:scene_ssim}

\paragraph{Spatio-temporal Signal Generation}
The frame-wise counting results are aggregated to create a temporal signal $\{\Delta C_1,\Delta C_2, ...,\Delta C_{T-1}\}$ where $\Delta C_i \vcentcolon= C_{i+1} - C_i$. This signal captures how the number of objects change throughout the video sequence.

\begin{figure}[!ht]
    \centering
    \subfloat[SSIM Frame 98]{\includegraphics[width=0.22\textwidth]{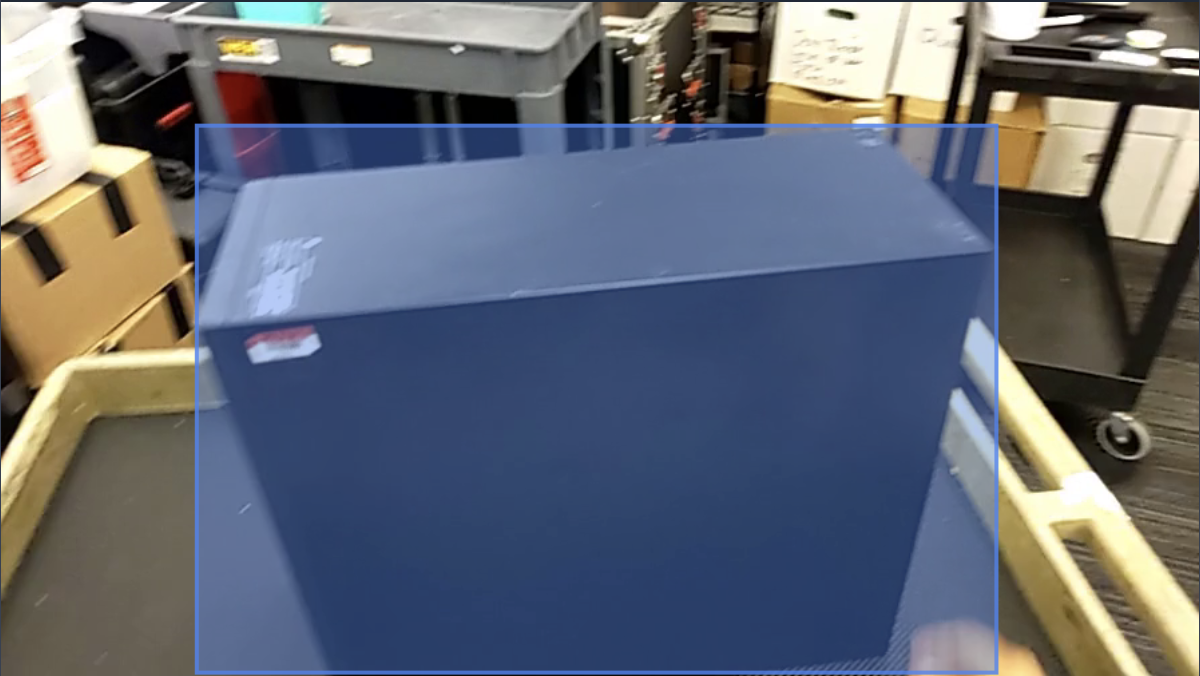}}
    \hfill
    \subfloat[SSIM Frame 99]{\includegraphics[width=0.22\textwidth]{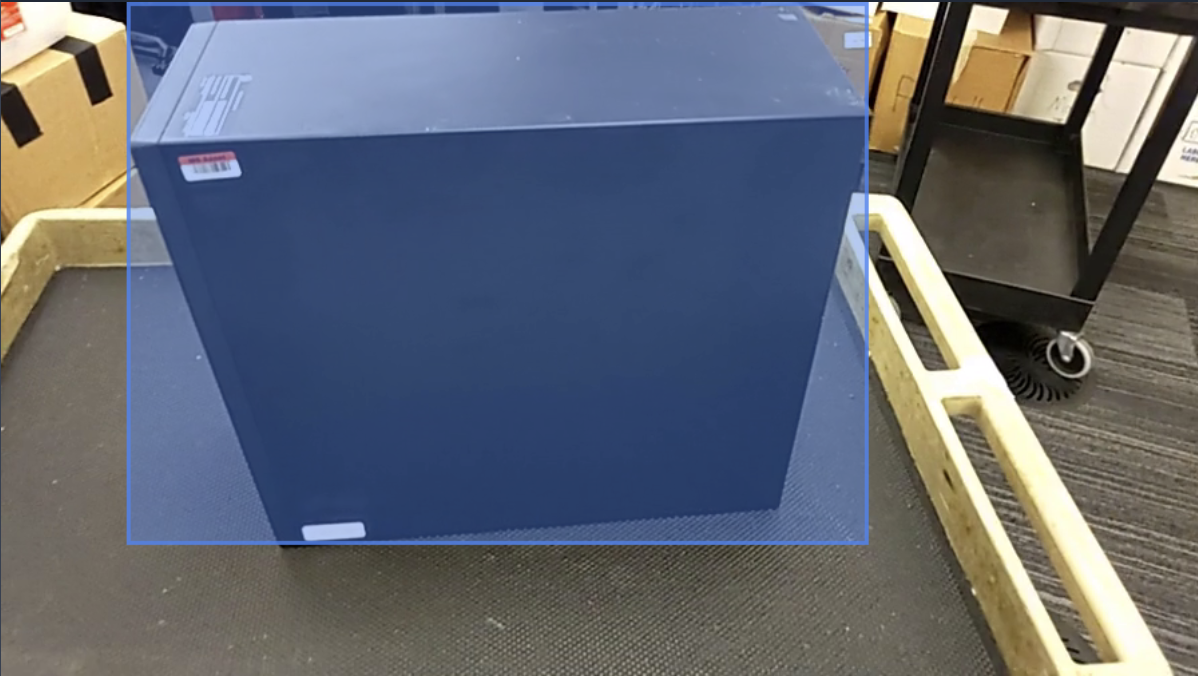}} \\
    \subfloat[SSIM Signal over entire video.]{\includegraphics[width=0.22\textwidth]{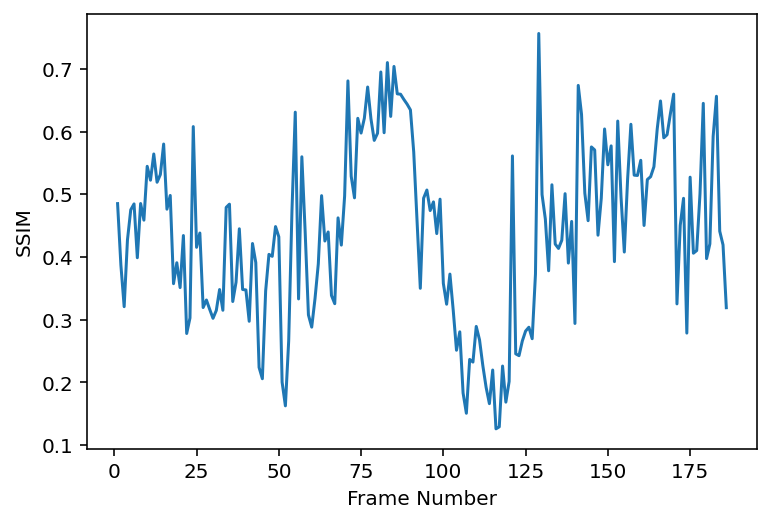}}
    \caption{Plot of SSIM filtering proactive interventions by expert agents in an image frame (time instance) of a video capturing a procedural task on how to assemble a RAM computer. Autonomous intervention happens at the 98th second in Frame 98}
    \label{fig:ssim_eg}
\end{figure}




To identify meaningful frames for proactive intervention, we employ the Structural Similarity Index Measure (SSIM)~\cite{wang2004image} to analyze temporal coherence between consecutive frames. This approach helps filter out redundant frames where the scene remains largely static, such as Figure \ref{fig:ssim_eg}, thus focusing interventions on moments of significant change.

Given two consecutive frames $f_i$ and $f_{i+1}$, we compute their SSIM as:
\begin{equation}
\label{eq:ssim}
    \text{SSIM}(f_i, f_{i+1}) = \frac{(2\mu_i\mu_{i+1} + c_1)(2\sigma_{i,i+1} + c_2)}{(\mu_i^2 + \mu_{i+1}^2 + c_1)(\sigma_i^2 + \sigma_{i+1}^2 + c_2)}
\end{equation}
where $\mu_i, \mu_{i+1}$ denote the mean intensities of frames $f_i$ and $f_{i+1}$, $\sigma_i^2, \sigma_{i+1}^2$ represent their respective variances, $\sigma_{i,i+1}$ is the covariance between the frames, $c_1 = (0.01L)^2$ and $c_2 = (0.03L)^2$ are stability constants, and $L = 255$ is the dynamic range of pixel values in grayscale.

The SSIM metric yields values in $[0,1]$, where higher values indicate greater structural similarity between frames. We leverage this property to identify and filter out frames with SSIM values exceeding a threshold $\tau$, effectively removing redundant temporal information. This filtering mechanism ensures that interventions are triggered only during meaningful scene changes, reducing unnecessary proactive interventions while maintaining responsiveness to significant environmental variations. Similarly to our temporal signal for changing object count, we consolidate the SSIM values for each frame into a set $\{s_1, s_2, ..., s_{T-1}\}$.

\subsection{Proactive Interactions and Interventions}

In the context of AI Agents, reactivity refers to the AI Agent's response to a user cue. In contrast, proactivity involves behaviors initiated by the AI Agent without user prompts. Proactive activities can be broadly categorized into proactive interaction and proactive intervention. Understanding the subtle but important distinction between these two is essential for leveraging the capabilities of a proactive agent. An AI Agent is considered to be \textit{proactively interacting} with the user if it initiates any engagement without user cues. Conversely, an AI Agent is \textit{proactively intervening} when it takes concrete steps to alter the user's behavior. By this definition, all proactive interventions are forms of proactive interactions, but not all proactive interactions qualify as interventions.

HoloAssist encompasses eight distinct categories of proactive behavior, of which three are classified as interventions, as detailed in Table~\ref{tab:interactions}.

\begin{table*}[!ht]
    \centering
    \begin{tabular}{p{0.25\textwidth} | c | c | p{0.45\textwidth}}
        \toprule
        \textbf{Conversation Type} & \textbf{Interaction} & \textbf{Intervention} & \textbf{Example} \\
        \midrule
        Follow-up Instruction & \checkmark & \checkmark & ``Put the battery back." \\ 
        Confirming Previous Action & \checkmark & \checkmark & ``Perfect." \\
        Correcting Wrong Action & \checkmark & \checkmark & ``Nope, not that one." \\
        Describe High-Level Instruction & \checkmark & $\times$ & ``You're going to validate, so we're going to move the shift lever through all of the settings." \\
        Opening Remarks & \checkmark & $\times$ & ``Now for this task, we are removing the graphic cards from the PC slot." \\ 
        Closing Remarks & \checkmark & $\times$ & ``You're all done. You can exit now." \\ 
        Adjusting Video & \checkmark & $\times$ & ``Just keep your eyes on your hands." \\ 
        Other & \checkmark & $\times$ & ``You can ground yourself again, but it's not really necessary." \\ 
        \bottomrule
    \end{tabular}
    \caption{Examples of Proactive Interactions and Proactive Interventions in HoloAssist.}
    \label{tab:interactions}
\end{table*}

\subsection{YETI Proactive Agent Intervention Algorithm}
\label{subsec:yeti_agent_algorithm}

The YETI algorithm incorporates several hyperparameters that determine when a Multimodal AI Agent should autonomously intervene proactively without any question or clarifications asked by the User Agent.


\begin{itemize}
    \item \textbf{SSIM threshold ($\tau$):} This parameter sets a filtering threshold. A frame's SSIM value with its corresponding frame must satisfy it to be considered in the YETI algorithm, to filter out highly similar frames where the user is not doing anything. In other words, if a frame and its proceeding frame have an SSIM of $\geq \tau$, the frame will not be considered for an autonomous intervention.
    
    \item \textbf{Conversation Interval ($m$):} This parameter enforces a minimum temporal gap between consecutive interventions, ensuring that the AI agent does not intervene too frequently. It defines the duration that must pass after an intervention before another can be initiated.
    
    \item \textbf{Local Extrema Range ($r$):} This parameter identifies the sensitivity of the algorithm to changes in object counts within the frames. It defines the range within which a change in object count must fall or rise to be considered significant.
    
    \item \textbf{Episode Interval ($k$):} This parameter limits the maximum rate at which interventions can occur by defining the length of an "episode." An episode is a consecutive sequence of frames within which only one intervention is permitted, thus preventing excessive and potentially disruptive interventions.
\end{itemize}

\begin{algorithm}[H]
\caption{YETI Proactive Intervention Detection}
\label{alg:intervention}
\begin{algorithmic}[1]
\Require{Frame sequence $F$ with alignment scores $\{\Delta C_i\}$}

\Ensure{Set of intervention frames $\mathcal{I}$}

\State Initialize empty sets: $\mathcal{I}$, $E_{obj}$ where $E_{obj}$ is the alignment scores per episode of frames 
\State Set episode interval $k$, conversation interval $m$, local extrema range $r$
\State Initialize episode count $n=0$, frame count $t=0$

\For{each frame $f_i \in F$}
    \If{$n > 0$ \textbf{and} $f_i \not\in$ conversation interval}
        \State $t \gets t + 1$
        \State Add $\Delta C_i$ to $E_{obj}$
        
        \If{$c_i \in$ local extrema range}
            \State $\mathcal{I} \gets \mathcal{I} \cup \{f_i\}$
        \EndIf
        
        \If{$t = k$}
            \State Update conversation interval
            \State Reset episode metrics
            \State $n \gets n + 1$, $t \gets 0$
        \EndIf
    \ElsIf{$n = 0$}
        \State $t \gets t + 1$
        \State Add $\Delta C_i$ to $E_{obj}$

        \If{$t = k$}
            \State $\Delta C_{min} \gets \min(E_{obj})$
            \State $\Delta C_{max} \gets \max(E_{obj})$
            \State Define local extrema range with tolerance $r$.
            \State $\mathcal{I} \gets \mathcal{I} \cup \{f_{current}\}$
            \State Update conversation interval
            \State Reset episode metrics
            \State $n \gets n + 1$, $t \gets 0$
        \EndIf
    \EndIf
\EndFor

\Return{$\mathcal{I}$}
\end{algorithmic}
\end{algorithm}

%% file: sec/4_experiments.tex
\section{Experiments}
\label{sec:experiments}

\begin{figure*}[htbp]
    \centering
    \subfloat[Intervention - 3 seconds]{
        \includegraphics[width=0.23\textwidth]{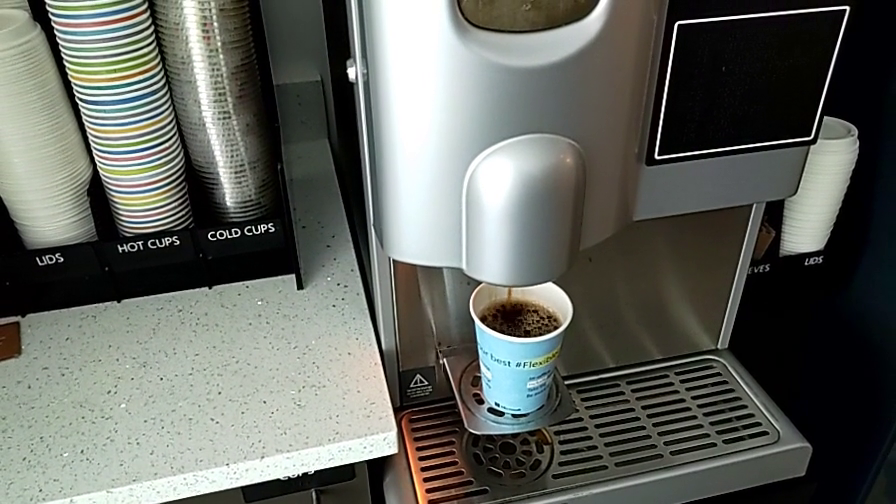}
        \label{fig:1}
    }
    \hfill
    \subfloat[Intervention - 2 seconds]{
        \includegraphics[width=0.23\textwidth]{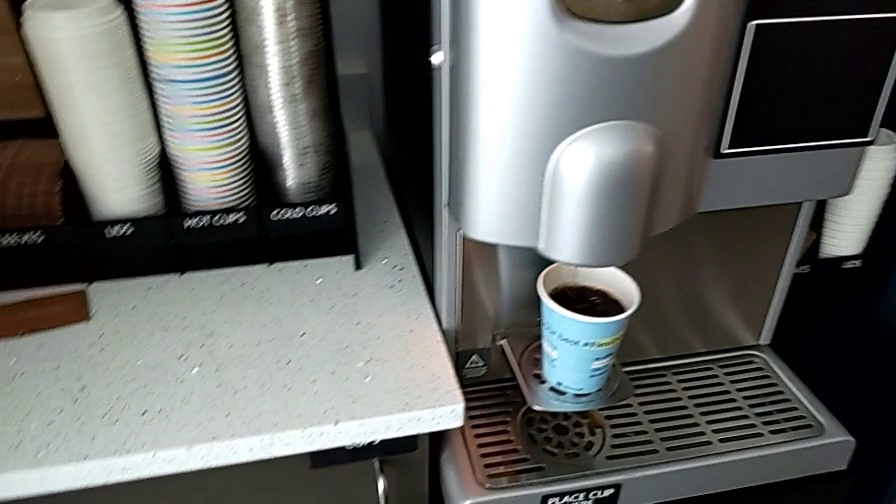}
        \label{fig:2}
    }
    \hfill
    \subfloat[Intervention - 1 second]{
        \includegraphics[width=0.23\textwidth]{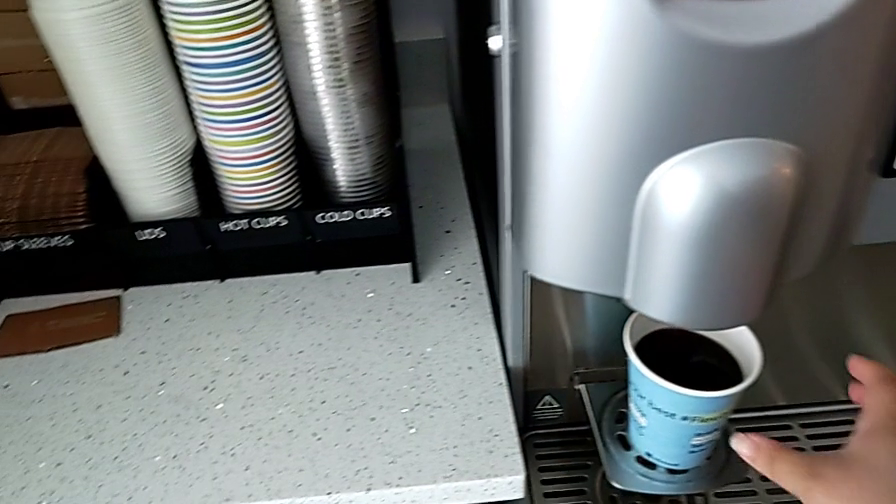}
        \label{fig:3}
    }
    \hfill
    \colorbox{green!30}{
    \subfloat[AI Agent proactively intervenes.]{
        \includegraphics[width=0.23\textwidth]{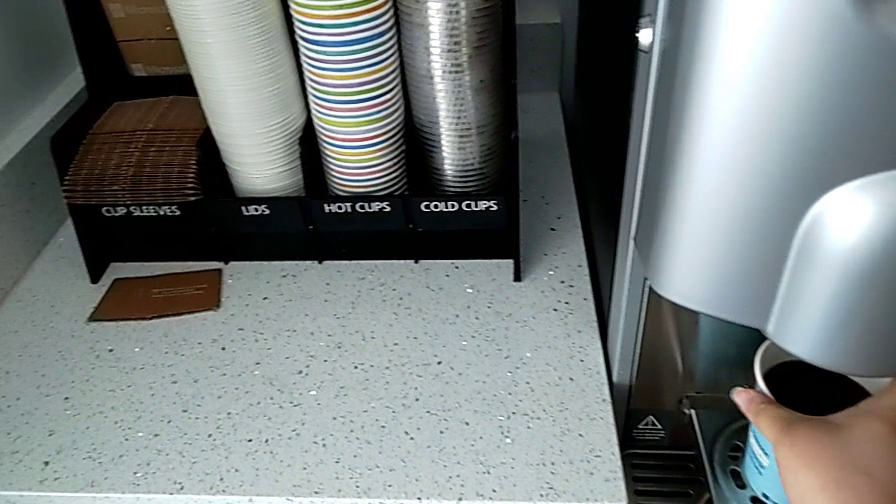}
        \label{fig:4}
    }
    }
    
    \vspace{1em}  
    
    \subfloat[SSIM between consecutive frames.]{
            \includegraphics[width=0.45\textwidth]{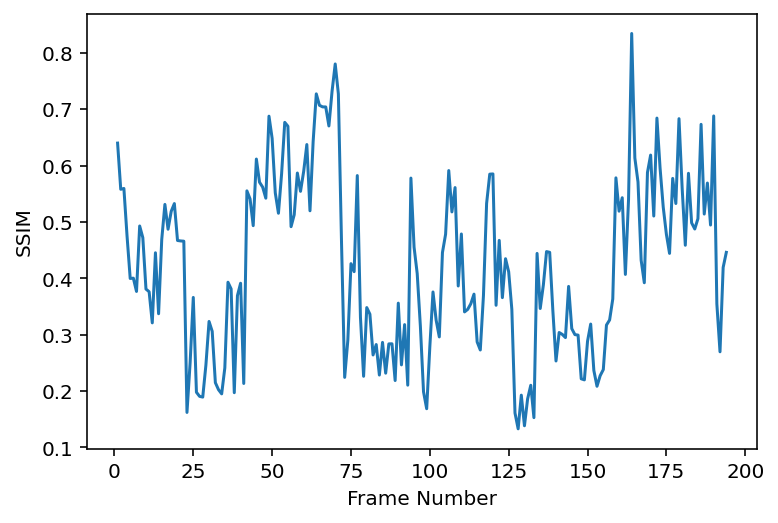}
        \label{fig:5}
    }
    \hfill
    \subfloat[Changing Objects Alignment Signal]{
        \includegraphics[width=0.45\textwidth]{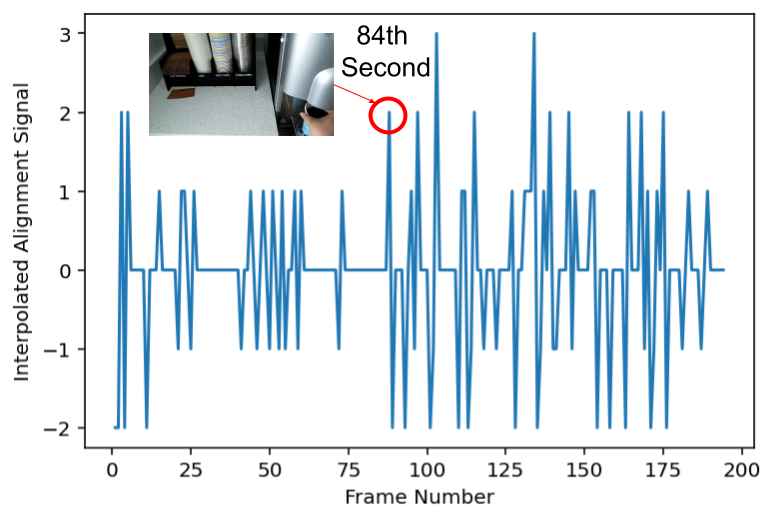} 
        \label{fig:6}
    }
    
    \caption{Intervention Detection for Coffee Making Task.}
    \label{fig:all}
\end{figure*}

To validate the results of our YETI algorithm, we conducted experiments with a wide variety of different settings. This also lets us see how each configuration of the algorithm and the value of each hyperparameter contributes to the evaluation metrics of our method compared to HoloAssist, the baseline for AI Agents proactively intervening with an user (student) task. An example of a proactive intervention being detected can be seen in Figure \ref{fig:all}.

\subsection{Experimental Settings}
\label{subsec:experimental_settings}

We carefully selected hyperparameters to balance the trade-off between timely interventions and avoiding excessive interruptions. The key parameters are summarized in Table~\ref{tab:hyperparameters}. A comprehensive analysis of hyperparameter sensitivity is provided in the supplementary material.

\begin{table}[!htbp]
    \centering
    \begin{tabular}{l|c}
    \toprule
    \textbf{Parameter} & \textbf{Value} \\
    \midrule
    SSIM threshold ($\tau$) & 0.9 \\
    Conversation interval ($m$) & 1 \\
    Extrema range ($r$) & $\pm$1 \\
    Minimum history ($k$) & 5 \\
    \bottomrule
    \end{tabular}
    \caption{Hyperparameters used in our experiments. $\tau$ controls frame similarity filtering, $m$ sets minimum gap between interventions, $r$ defines the range for local minima detection, and $k$ specifies required history length before intervention.}
    \label{tab:hyperparameters}
    \vspace{-10pt}
\end{table}

We evaluate two variants of our YETI algorithm:
\begin{itemize}
    \item \textbf{Global YETI}: Uses the first detected local extrema as a fixed threshold throughout the sequence
    \item \textbf{Local YETI}: Continuously updates the extrema threshold based on recent history
\end{itemize}

This allows us to analyze the impact of adaptive versus fixed intervention thresholds on system performance. The full results of evaluating performance in this way can be seen in Tables~\ref{tab:intervention-results} and ~\ref{tab:interaction-results}. 


For the comparative analysis between our YETI methods and HoloAssist, it is important to note we limited our evaluation to videos where the user and the expert agent both initiated conversations. This stipulation more closely aligns with the real-world assistive AI Agent scenario we had in mind, where the user may ask for help completing a task. This turns out to be 482 videos out of the total HoloAssist dataset, more than enough to be a representative sample. We also aggregated the HoloAssist results for each intervention class into one through averaging in order to have a head-to-head comparison with regards to predicting any kind of intervention. We also provide Supplemental results where the user agent does not communicate with any dialogue, however the expert agent proactively interacts and intervenes with the user, based on user's visual observations.

\begin{table*}[]
    \centering
    \setlength{\tabcolsep}{4pt}
    \begin{tabular}{c|c|ccc|ccc|ccc|ccc}
    \toprule
    & \multirow{2}{*}{Method} & \multicolumn{3}{c}{Overall} & \multicolumn{3}{|c}{Confirm Action} & \multicolumn{3}{|c}{Correct Mistake} & \multicolumn{3}{|c}{Follow Up} \\
    &   & Prec. & Rec. & F-meas. & Prec. & Rec. & F-meas. & Prec. & Rec. & F-meas. & Prec. & Rec. & F-meas. \\
    \midrule
    \multirow{5}{*}{HoloAssist} 
    & (RGB) & 13.93 & 33.33 & 19.65 & 0.00 & 0.00 & 0.00 & 0.00 & 0.00 & 0.00 & 41.79 & 100.00 & 58.95 \\
    & (R+H) & 24.89 & 33.64 & 28.61 & 32.14 & 4.50 & 7.89 & 0.00 & 0.00 & 0.00 & 42.52 & 96.43 & 59.02 \\
    & (R+E) & 25.55 & 33.73 & 29.08 & 33.90 & 10.36 & 15.87 & 0.00 & 0.00 & 0.00 & 42.76 & 90.83 & 58.15 \\
    & (R+H+E) & 48.31 & 37.59 & 42.28 & 39.11 & 40.93 & 40.00 & 61.11 & 9.91 & 17.05 & 44.70 & 61.93 & 51.92 \\
    & (RGB[Pt]) & 37.54 & 37.74 & 37.64 & 42.31 & 27.50 & 33.33 & 27.33 & 36.61 & 31.30 & 42.97 & 49.11 & 45.84 \\
    \midrule
    \multirow{2}{*}{YETI (Ours)}
    & Global & 41.86 & 88.31 & 56.17 & 22.54 & 91.51 & 36.17 & 11.68 & 90.29 & 20.69 & 27.55 & 89.69 & 42.15\\
    & Local & 46.88 & 60.38 & 52.77 & 26.55 & 68.62 & 38.29 & 14.71 & 68.02 & 24.18 & 30.07 & 62.05 & 40.51 \\
    \bottomrule
    \end{tabular}
    \caption{Performance of YETI compared to HoloAssist in detecting proactive interventions.}
    \label{tab:intervention-results}
\end{table*}


\begin{table}[]
  \centering
  \setlength{\tabcolsep}{4pt}
  \begin{tabular}{c|c|cccc}
  \toprule
  & \multirow{2}{*}{Method} & \multicolumn{4}{c}{Overall} \\
  &   & Acc. & Prec. & Rec. & F1 \\
  \midrule
  \multirow{2}{*}{YETI (Ours)}
  & Global & 86.97 & 52.23 & 87.04 & 65.28 \\
  & Local & 93.76 & 64.51 & 59.73 & 62.02 \\
  \bottomrule
  \end{tabular}
  \caption{Evaluation of YETI on our proactive interaction detection setting which encompasses the proactive intervention benchmark}
  \label{tab:interaction-results}
\end{table}

\subsection{Metrics}
\label{subsec:metrics}

We evaluate our YETI algorithm's effectiveness in detecting appropriate moments for proactive interaction and intervention using standard classification metrics:

\begin{equation}
\text{Accuracy} = \frac{\text{TP} + \text{TN}}{\text{TP} + \text{TN} + \text{FP} + \text{FN}}
\label{eq:accuracy}
\end{equation}

\begin{equation}
\text{Precision} = \frac{\text{TP}}{\text{TP} + \text{FP}}
\label{eq:precision}
\end{equation}

\begin{equation}
\text{Recall} = \frac{\text{TP}}{\text{TP} + \text{FN}}
\label{eq:recall}
\end{equation}

\begin{equation}
\text{F-measure} = 2 \cdot \frac{\text{Precision} \cdot \text{Recall}}{\text{Precision} + \text{Recall}}
\label{eq:fscore}
\end{equation}

\noindent where TP, TN, FP, and FN denote true positives, true negatives, false positives, and false negatives, respectively. These metrics provide a comprehensive assessment of YETI's intervention capabilities, measuring both its ability to intervene at appropriate moments and its capacity to avoid unnecessary interruptions.

As there is no empiric measurement of when the ``best" precise moment to intervene is, we use a window of five seconds around HoloAssist labelled ground-truth proactive intervention starting time-stamps in order to assess whether a detected intervention frame is a true positive or a false positive. This is consistent with the method used to evaluate the HoloAssist baseline model~\cite{wang2023holoassist}, which also determines true positives and false positives by measuring temporal proximity to the nearest labelled intervention and seeing if it is within a window of tolerance.








\subsection{Results}
\label{subsec:results}

We can see in Tables \ref{tab:interaction-results} and \ref{alg:intervention} that YETI achieves impressive results when detecting proactive interventions and detections. Furthermore, as PaliGemma's scale is on the order of three billion parameters, it is suitable to be used on board an Augmented Reality device. The small data size of SSIM and Object Count Alignment relative to other features, as seen in Table \ref{table:data_sizes}, means that AI Agents on the device can use the information while having a small response time.

%% file: sec/5_conclusion.tex
\section{Conclusion and Future Work}
\label{sec:conclusion_future}


In this work, we tackle the significant gap in proactive AI assistance for everyday real-world tasks by introducing the Yet-to-Intervene (YETI) algorithm. Traditional AI agents, such as those exemplified by the HoloAssist baseline, predominantly operate reactively, responding only to explicit user prompts. This reactive nature limits their effectiveness in dynamic and context-sensitive environments where anticipatory intervention can substantially enhance user experience and task efficiency. Our YETI algorithm addresses this limitation by enabling AI agents to proactively identify and intervene in user actions without awaiting explicit cues.

Integrating YETI with lightweight Vision-Language Models (VLMs) like PaliGemma, our approach demonstrates remarkable improvements across key performance metrics, including Recall, Precision, Accuracy, and F-Measure. Notably, YETI achieves these enhancements while utilizing features that are up to 60,000 times more memory-efficient than those employed by state-of-the-art models such as HoloAssist. This drastic reduction in computational overhead not only makes YETI more accessible for deployment on resource-constrained devices but also paves the way for real-time, scalable AI assistance in diverse augmented reality (AR) applications.

Our extensive ablation studies in the paper and supplemental reveal the critical features that significantly influence the accurate detection of intervention moments. By systematically varying and analyzing different feature sets, we identify the most impactful components that empower YETI to discern when proactive interactions or interventions are warranted. This insight underscores the importance of feature selection in the design of efficient and effective AI assistance algorithms.

Looking ahead, there are several promising avenues for future research. First, we aim to enhance YETI's intervention capabilities by incorporating richer sensory data, including hand pose, eye gaze, head orientation, Inertial Measurement Unit (IMU) readings, and depth information. Integrating these modalities is expected to provide a more comprehensive understanding of the user's context and intentions, thereby enabling more nuanced and timely interventions. Second, we plan to evaluate YETI's performance across a broader spectrum of VLMs to assess its generalizability and identify optimal model architectures for proactive assistance. Additionally, fine-tuning VLMs on the HoloAssist dataset could further refine the agent's ability to anticipate user needs and improve intervention accuracy.

Furthermore, future implementations of YETI will explore adaptive learning mechanisms that allow the AI agent to continuously refine its intervention strategies based on user feedback and evolving task dynamics. This adaptive approach is anticipated to enhance the personalization and effectiveness of AI assistance, making it more attuned to individual user preferences and behaviors.

In summary, the YETI algorithm represents a significant advancement in the development of proactive AI assistants, offering enhanced performance with reduced computational demands. By enabling timely and context-aware interventions, YETI has the potential to transform human-AI collaboration in AR environments and beyond. Our ongoing and future work seeks to build upon these foundations, further pushing the boundaries of what proactive AI agents can achieve in facilitating and automating complex real-world tasks.

%% file: sec/act_supplementary.tex
\clearpage
\setcounter{page}{1}
\maketitlesupplementary


\section{Detailed Analysis of Results}
\label{sec:sup-results}

This section provides a more in-depth analysis of the results presented in Tables \ref{table:data_sizes}, \ref{tab:intervention-results}, and \ref{tab:interaction-results} with a special focus on Table \ref{tab:intervention-results} comparing the performance of YETI to the HoloAssist baseline in detecting proactive interventions.

\textbf{Overall Performance:}

YETI consistently outperforms HoloAssist \cite{wang2023holoassist} overall, demonstrating a significant improvement in accurately detecting proactive interventions in most scenarios. This superior performance is observed in both the Global and Local variants of YETI. Notably, YETI achieves substantially higher recall in all categories, indicating its effectiveness in identifying a greater proportion of actual proactive interventions. This improvement is crucial for ensuring that the AI agent can effectively assist users by recognizing and responding to their needs in a timely manner.

\textbf{Specific Intervention Types:}

\begin{itemize}
    \item \textbf{Overall:}
    When taking into account all intervention types, both Local and Global variants of YETI substantially outperform the HoloAssist baseline on recall and F-measure, while surpassing precision in every modality of HoloAssist except R+H+E (RGB video + Hand pose + Eye gaze), where it still remains competitive.
    \item \textbf{Confirm Action:}
    In only considering interventions where the AI Agent should step in to confirm the action the user is taking, YETI continues to achieve substantially higher recall than HoloAssist. The F-measure, which seeks to balance out precision and recall, is also higher than every combination of modalities used for HoloAssist besides R+H+E.
    \item \textbf{Correct Mistake:}
    While both YETI and HoloAssist struggle to detect when to intervene to correct a mistake by the user, many modalities for HoloAssist fail to detect them entirely, and for the ones that can detect some of these interventions, the recall is very poor. YETI, on the other hand, has extremely high recall but comparatively poor precision. This contributes to YETI having a higher F-measure than all HoloAssist modalities besides HoloAssist RGB[Pt] (model pretrained on ImageNet).
    \item \textbf{Follow Up:}
    Following up with the user after the user performs an action is where YETI struggles the most compared to the HoloAssist baselines, but it still remains competitive, outperforming the R+H+E and RGB[Pt] baselines on recall. Ironically, even though YETI and R+H+E and RGB[Pt] performed better on every other category, they had the worst results in following up. This indicates that more sophisticated features may complicate the process of detecting Follow-up interventions.
\end{itemize}

\textbf{Further Comparison with HoloAssist:}

The performance gains of YETI over HoloAssist can be attributed to several factors:

\begin{itemize}
    \item \textbf{Global / Local Context:} YETI's ability to leverage its history contributes to its improved accuracy in detecting proactive interventions. By considering both the broader context of the interaction, as in the case of Global YETI, and the specific details of the current situation, as in the case of local YETI, YETI can better understand the user's intentions and needs.
    \item \textbf{Advanced Feature Representation:} YETI likely employs more streamlined feature representations compared to HoloAssist, allowing it to capture more nuanced aspects of the interaction and make more informed decisions about potential interventions.
    \item \textbf{Data Efficiency:} The smaller data size of SSIM and Object Count Alignment features, as highlighted in Table \ref{table:data_sizes}, enables YETI to process information and respond to user actions with minimal latency. This efficiency is crucial for real-time applications in augmented reality environments.
\end{itemize}

YETI's superior performance in detecting proactive interventions, coupled with its efficiency and suitability for deployment on augmented reality devices, makes it a promising approach for enhancing human-computer interaction in various domains. Its ability to accurately detect when to proactively intervene has the potential to significantly improve user experiences and facilitate more effective collaboration between humans and AI agents.

\section{Comparative Analysis of YETI}

\subsection{Comparison with other Classifier Models}

In order to assess the efficacy of our YETI algorithm (Algorithm \ref{alg:intervention}), we trained a Random Forest Classifier, a Decision Tree, and a Multi-Layer Perceptron (MLP) in order to have a comparative analysis. Statistics about the data used to train and evaluate the Random Forest Classifier can be seen in Table \ref{tab:supp-data-splits}.
Due to the imbalance in labels, it was more statistically safe for the models to assume that every single frame was a negative example of a proactive interaction or a proactive intervention, resulting in accuracy being high but precision, recall, and F-measure being zero. This justifies the use of our algorithm to detect when to intervene rather than relying solely on a model such as these ones, which are sensitive to skewed data distributions. 


\begin{table*}[] 
  \centering
  \setlength{\tabcolsep}{4pt}
  \begin{tabular}{c|c|c|c|c|c|c}
  \toprule
   & Interactions & Interventions & Confirm Action & Correct Mistake & Follow Up & Total \\
  \midrule
  
  Train & 7285 (6.77\%) & 4562 (4.24\%) & 2222 (2.07\%) & 526 (0.49\%)  & 1814 (1.37 \%) & 107592 (81.43\%) \\
  \midrule
  Test & 1583 (6.45\%) & 1005 (4.1\%) & 474 (1.93\%) & 142 (0.58\%) & 389 (1.59\%) & 24537 (18.57\%) \\
  \midrule
  Total & 8868 (6.72\%) & 5567 (4.21\%) & 2696 (2.04\%) & 668 (0.51\%) & 2203 (1.67\%) & 132129\\
  \bottomrule
  \end{tabular}
  \caption{Training and Test splits of Image Frames with Distributions of Proactive Interactions, Proactive Interventions and 3 Proactive Intervention types}
  \label{tab:supp-data-splits}
\end{table*}

\begin{table*}[!ht]
  \centering
  \setlength{\tabcolsep}{4pt}
  \begin{tabular}{c|c|c|c|c|c}
  \toprule
   & Interactions & Interventions & Confirm Action & Correct Mistake & Follow Up \\
  \midrule
  Random Forest Classifier & & & & & \\
  Decision Tree Classifier & 93.55 & 95.9 & 98.41 & 99.42 & 98.07 \\
  MLP Classifier & & & & & \\
  \midrule
  Global YETI (Ours) & 86.97 & 84.85 & 80.75 & 79.97 & 82.36 \\
  \midrule
  Local YETI (Ours) & 93.76 & 93.36 & 93.05 & 93.32 & 93.31 \\
  \bottomrule
  \end{tabular}
  \caption{Accuracies of different classification models in proactivity prediction. Random Forest (RF), Decision Tree (DT) and Multi-Layer Perceptron (MLP) Classifiers have high accuracies as they can predict when the AI Agent should not be proactively interacting or intervening. The same trend is observed in the 3 different proactive intervention type predictions like confirming actions, correcting mistakes, following up instructions. However, RF, DT and MLP classifiers cannot classify any true positive (actual proactive instances) due to the extremely sparse and skewed distribution of proactivity labels in Table \ref{tab:supp-data-splits}, thus has 0 precision and recall. These classifiers cannot detect proactivity on-the-fly in real time. Our YETI Proactive Agent Detection Algorithm has good precision and high recall model architectures in detecting proactivity on-the-fly as shown in Table \ref{tab:intervention-results}.}
  \label{tab:supp-data-splits}
\end{table*}

\subsection{Comparison with Implicit Expert-User Agent Setting}

In Table \ref{tab:intervention-results}, we only presented results using data from HoloAssist examples with both student-led and instructor-led conversations. In Tables \ref{tab:interventionssupp} and \ref{tab:interactionssupp} we present the results for the 1191 HoloAssist examples where there was not a student-led conversation, instead relying on implicit student-instructor interactions where, for example, the instructor tells the student to do something and the student complies without further comment.

\begin{table*}[]
    \centering
    \setlength{\tabcolsep}{4pt}
    \begin{tabular}{c|c|ccc|ccc|ccc|ccc}
    \toprule
    & \multirow{2}{*}{Method} & \multicolumn{3}{c}{Overall} & \multicolumn{3}{|c}{Confirm Action} & \multicolumn{3}{|c}{Correct Mistake} & \multicolumn{3}{|c}{Follow Up} \\
    &   & Prec. & Rec. & F-meas. & Prec. & Rec. & F-meas. & Prec. & Rec. & F-meas. & Prec. & Rec. & F-meas. \\
    \midrule
    \multirow{5}{*}{HoloAssist} 
    & (RGB) & 13.93 & 33.33 & 19.65 & 0.00 & 0.00 & 0.00 & 0.00 & 0.00 & 0.00 & 41.79 & 100.00 & 58.95 \\
    & (R+H) & 24.89 & 33.64 & 28.61 & 32.14 & 4.50 & 7.89 & 0.00 & 0.00 & 0.00 & 42.52 & 96.43 & 59.02 \\
    & (R+E) & 25.55 & 33.73 & 29.08 & 33.90 & 10.36 & 15.87 & 0.00 & 0.00 & 0.00 & 42.76 & 90.83 & 58.15 \\
    & (R+H+E) & 48.31 & 37.59 & 42.28 & 39.11 & 40.93 & 40.00 & 61.11 & 9.91 & 17.05 & 44.70 & 61.93 & 51.92 \\
    & (RGB[Pt]) & 37.54 & 37.74 & 37.64 & 42.31 & 27.50 & 33.33 & 27.33 & 36.61 & 31.30 & 42.97 & 49.11 & 45.84 \\
    \midrule
    \multirow{2}{*}{YETI (Ours)}
    & Global & 41.86 & 88.31 & 56.17 & 22.54 & 91.51 & 36.17 & 11.68 & 90.29 & 20.69 & 27.55 & 89.69 & 42.15\\
    & Local & 46.88 & 60.38 & 52.77 & 26.55 & 68.62 & 38.29 & 14.71 & 68.02 & 24.18 & 30.07 & 62.05 & 40.51 \\
    \midrule
    \multirow{2}{*}{Implicit}
    & Global & 28.96 & 92.59 & 44.11 & 15.68 & 94.76 & 26.90 & 11.28 & 93.58 & 20.12 & 20.47 & 92.56 & 33.53 \\
    & Local & 34.55 & 71.71 & 46.63 & 18.4 & 77.71 & 29.76 & 14.28 & 74.21 & 23.95 & 24.51 & 72.48 & 36.64 \\
    \bottomrule
    \end{tabular}
    \caption{Evaluation of YETI compared to HoloAssist and Implicit Interventions}
    \label{tab:interventionssupp}
\end{table*}

\begin{table}[]
  \centering
  \setlength{\tabcolsep}{4pt}
  \begin{tabular}{c|c|cccc}
  \toprule
  & \multirow{2}{*}{Method} & \multicolumn{4}{c}{Overall} \\
  &   & Acc. & Prec. & Rec. & F1 \\
  \midrule
  \multirow{2}{*}{YETI (Ours)}
  & Global & 86.97 & 52.23 & 87.04 & 65.28 \\
  & Local & 93.76 & 64.51 & 59.73 & 62.02 \\
  \midrule
  \multirow{2}{*}{Implicit}
  & Global & 81.37 & 35.68 & 91.9 & 51.42 \\
  & Local & 92.62 & 45.67 & 71.28 & 55.67 \\
  \bottomrule
  \end{tabular}
  \caption{Evaluation of YETI compared to HoloAssist and Implicit Interactions}
  \label{tab:interactionssupp}
\end{table}

\section{Additional Related Works}

\subsection{Procedural Mistake Detection}

Previous studies have explored the analysis of egocentric data to assist with procedural tasks; however, none have approached this challenge in the same comprehensive manner as YETI. YETI is designed to detect optimal moments for proactive intervention. In contrast, existing works primarily focus on mistake detection, which limits direct comparisons with YETI’s broader intervention detection capabilities.

\textbf{PREGO} \cite{flaborea2024prego} (Mistake Detection in \textbf{PR}ocedural \textbf{EGO}centric Videos) targets online mistake detection in a manner akin to YETI's real-time intervention detection. Nevertheless, PREGO is confined to the Mistake Detection intervention type and does not address other scenarios where AI intervention could be beneficial. Specifically, PREGO lacks mechanisms for proactive interventions, requiring users to make errors before the AI can respond. Although PREGO incorporates step anticipation, it only detects deviations from predefined plans to identify mistakes. Additionally, PREGO necessitates a symbolic description of the task for mistake identification, whereas YETI operates directly on video frames without such annotations.

\textbf{TI-PREGO} \cite{plini2024ti} extends the work of PREGO with a more comprehensive integration of Large Language Models (LLMs) for action anticipation and detection modules, incorporating chain-of-thought reasoning and in-context learning. They evaluate the ability of several LLaMA, Mistral, Gemma, and GPT models to perform step anticipation. Similar to PREGO, TI-PREGO remains focused solely on mistake detection within procedural tasks and does not expand into proactive intervention detection.

\textbf{Differentiable Task Graph Learning} \cite{seminara2024differentiable} also addresses online mistake detection but diverges by utilizing Task Graphs instead of LLM-based symbolic reasoning. Task Graphs model procedures as sequences of steps with directed dependencies, ensuring certain steps precede others. However, this approach requires pre-segmented key-step sequences from input videos, rendering it unsuitable for real-time Augmented Reality applications. In contrast, YETI processes continuous actions without necessitating pre-annotated data, allowing for real-time operation.

\textbf{Eyes Wide Unshut} \cite{mazzamuto2024eyes} mitigates the reliance on supervised learning observed in Differentiable Task Graph Learning by predicting mistakes based on eye gaze trajectories instead of manual annotations. This method forecasts the user's next gaze position during task execution and compares it with actual gaze data to detect discrepancies. Unlike YETI, Eyes Wide Unshut heavily depends on the availability of eye gaze data, which may not always be accessible. Furthermore, many tasks do not require significant gaze shifts, limiting the method's applicability and increasing the potential for false positives when gaze changes are unrelated to task performance errors.

\subsection{Action Detection}

Other works have also explored broad action detection, classifying actions detected in videos instead of detecting when to proactively intervene or when a user makes a mistake.

\textbf{Quasi-Online Detection of Take and Release Actions} \cite{scavo2023quasi} focuses on near real-time detection, allowing a slight delay between an action occurring and its detection. This work concentrates on identifying "take" and "release" actions—instances where the user's hands interact with objects—rather than on mistake or intervention detection. While such action detection could support downstream tasks like intervention or mistake identification, it does not directly address these areas, thereby limiting its overall scope and applicability for use as an AI assistant.

\textbf{OadTR} \cite{wang2021oadtr} is one of the first works to use Transformer-based models for online action detection, pivoting from Recurrent Neural Networks (RNNs), which exhibit less parallelism. Unlike the other works in this list, however, the video in the datasets used to evaluate and OadTR are not egocentric, so it is not clear whether the models would generalize to detect actions from video with a different perspective.

\section{Ablation Studies}

\subsection{Agent Conversation Interval}

The Agent Conversation Interval is a parameter for how long we suppose it will take a user to respond to an intervention by the AI Agent. In Table \ref{tab:hyperparameters} we use a value of one, indicating the user will take about one second to act on the advice of the proactive AI Agent. Here we explore alternative intervals where a user might respond to the intervention within up to five seconds rather than one second.

Table \ref{tab:interventionssuppconvnit} shows precision, recall, and F-measure for detecting each kind of proactive intervention, with varying values for the conversation interval. We see that as we increase the conversation interval, the F-score decreases, showing that overall, a conversation interval of one is the best choice. The same logic follows for detection proactive interactions, shown in Table \ref{tab:supp-agent-conv-interactions}, where we also see that F-measure decreases as the conversation interval increases. 

\begin{table*}[!ht]
    \centering
    \setlength{\tabcolsep}{4pt}
    \begin{tabular}{c|c|ccc|ccc|ccc|ccc}
    \toprule
    \multirow{2}{*}{Method} & \multirow{2}{*}{Conv. Interval} & \multicolumn{3}{c}{Overall} & \multicolumn{3}{|c}{Confirm Action} & \multicolumn{3}{|c}{Correct Mistake} & \multicolumn{3}{|c}{Follow Up} \\
     & & Prec. & Rec. & F-meas. & Prec. & Rec. & F-meas. & Prec. & Rec. & F-meas. & Prec. & Rec. & F-meas. \\
    \midrule
    \multirow{5}{*}{Global YETI}
     & 1 & 41.86 & 88.31 & 56.17 & 22.54 & 91.51 & 36.17 & 11.68 & 90.29 & 20.69 & 27.55 & 89.69 & 42.15\\
     & 2 & 41.06 & 84.09 & 55.18 & 22.52 & 88.33 & 35.89 & 11.49 & 86.52 & 20.29 & 27.44 & 85.87 & 41.59\\
     & 3 & 41.13 & 80.49 & 54.44 & 22.41 & 85.42 & 35.50 & 11.50 & 83.28 & 20.22 & 27.31 & 82.53 & 41.04\\
     & 4 & 41.08 & 77.09 & 53.61 & 22.53 & 82.81 & 35.43 & 11.54 & 80.46 & 20.19 & 27.33 & 79.34 & 40.65\\
     & 5 & 40.95 & 73.91 & 52.70 & 22.31 & 80.03 & 34.89 & 11.59 & 77.92 & 20.19 & 26.96 & 76.22 & 39.83\\
     \midrule
     \multirow{5}{*}{Local YETI}
    & 1 & 46.88 & 60.38 & 52.77 & 26.55 & 68.62 & 38.29 & 14.71 & 68.02 & 24.18 & 30.07 & 62.05 & 40.51\\
     & 2 & 47.01 & 56.4 & 51.27 & 26.54 & 64.97 & 37.69 & 13.92 & 63.37 & 22.82 & 30.39 & 58.54 & 40.01\\
     & 3 & 47.15 & 56.39 & 51.36 & 25.93 & 64.46 & 36.99 & 14.35 & 63.52 & 23.41 & 30.75 & 59.26 & 40.49\\
     & 4 & 46.43 & 53.66 & 49.78 & 25.78 & 62.39 & 36.49 & 13.32 & 58.98 & 21.73 & 30.41 & 56.45 & 39.53\\
     & 5 & 45.81 & 51.01 & 48.27 & 26.05 & 60.58 & 36.44 & 13.37 & 56.71 & 21.64 & 29.47 & 52.99 & 37.88\\
    \bottomrule
    \end{tabular}
    \caption{Evaluation of YETI for Agent Conversation Intervals for proactive interventions}
    \label{tab:interventionssuppconvnit}
\end{table*}

\begin{table}[!ht]
  \centering
  \setlength{\tabcolsep}{4pt}
  \begin{tabular}{c|c|cccc}
  \toprule
  \multirow{2}{*}{Method} & \multirow{2}{*}{Conv. Interval} & \multicolumn{4}{c}{Overall} \\
  &   & Acc. & Prec. & Rec. & F1 \\
  \midrule
  \multirow{5}{*}{Global YETI}
  & 1 & 86.97 & 52.23 & 87.04 & 65.28 \\
  & 2 & 90.29 & 52.30 & 82.51 & 64.02 \\
  & 3 & 92.01 & 52.42 & 78.63 & 62.90 \\
  & 4 & 93.11 & 52.69 & 75.20 & 61.97 \\
  & 5 & 92.65 & 51.15 & 76.22 & 61.22 \\
  \midrule
  \multirow{5}{*}{Local YETI}
  & 1 & 93.76 & 64.51 & 59.73 & 62.02 \\
  & 2 & 94.25 & 63.79 & 55.47 & 59.34 \\
  & 3 & 94.59 & 63.43 & 55.12 & 58.98 \\
  & 4 & 94.77 & 62.23 & 52.3 & 56.84 \\
  & 5 & 94.98 & 61.61 & 49.79 & 55.08 \\
  \bottomrule
  \end{tabular}
  \caption{Comparison over different Agent Conversation interval sizes for proactive interactions.}
  \label{tab:supp-agent-conv-interactions}
\end{table}

\subsection{Extrema Range}

The extrema range refers to how close the alignment score needs to be to a minima or maxima (whether local or global) in order for a proactive intervention to be detected. A larger extrema range makes YETI more sensitive, resulting in more proactive detections, while a narrower extrema range leads to less proactive detections.
We see in Tables \ref{tab:supp-extrema-interventions} and \ref{tab:supp-extrema-interactions} that an extrema range of zero results in a much lower recall, while an extrema range of two results in a lower precision. Since overall, YETI struggles more with precision than recall, our chosen extrema range is one, which maximizes precision.

\begin{table*}[!ht]
    \centering
    \setlength{\tabcolsep}{4pt}
    \begin{tabular}{c|c|ccc|ccc|ccc|ccc}
    \toprule
    \multirow{2}{*}{Method} & \multirow{2}{*}{Extrema Ranges} & \multicolumn{3}{c}{Overall} & \multicolumn{3}{|c}{Confirm Action} & \multicolumn{3}{|c}{Correct Mistake} & \multicolumn{3}{|c}{Follow Up} \\
     & & Prec. & Rec. & F-meas. & Prec. & Rec. & F-meas. & Prec. & Rec. & F-meas. & Prec. & Rec. & F-meas. \\
    \midrule
    \multirow{3}{*}{Global YETI}
     & 0 & 41.34 & 70.61 & 52.15 & 22.52 & 76.96 & 34.85 & 11.09 & 73.87 & 19.28 & 27.13 & 73.92 & 39.69\\
     & $\pm1$ & 41.86 & 88.31 & 56.17 & 22.54 & 91.51 & 36.17 & 11.68 & 90.29 & 20.69 & 27.55 & 89.69 & 42.15\\
     & $\pm2$ & 41.05 & 91.95 & 56.75 & 22.19 & 94.09 & 35.91 & 11.48 & 93.08 & 20.44 & 27.43 & 93.04 & 42.37\\
     \midrule
     \multirow{3}{*}{Local YETI}
     & 0 & 43.42 & 26.02 & 32.54 & 24.83 & 33.16 & 28.39 & 11.12 & 28.28 & 15.96 & 27.35 & 27.17 & 27.26\\
     & $\pm1$ & 46.88 & 60.38 & 52.77 & 26.55 & 68.62 & 38.29 & 14.71 & 68.02 & 24.18 & 30.07 & 62.05 & 40.51\\
     & $\pm2$ & 46.74 & 77.23 & 58.24 & 25.98 & 83.11 & 39.59 & 13.34 & 79.83 & 28.86 & 30.63 & 79.22 & 44.18\\
    \bottomrule
    \end{tabular}
    \caption{Evaluation of YETI for Extrema Ranges for proactive interventions}
    \label{tab:supp-extrema-interventions}
\end{table*}

\begin{table}[!ht]
  \centering
  \setlength{\tabcolsep}{4pt}
  \begin{tabular}{c|c|cccc}
  \toprule
  \multirow{2}{*}{Method} & \multirow{2}{*}{Extrema Ranges} & \multicolumn{4}{c}{Overall} \\
  &   & Acc. & Prec. & Rec. & F1 \\
  \midrule
  \multirow{3}{*}{Global YETI}
  & 0 & 92.14 & 53.94 & 68.52 & 60.36 \\
  & $\pm1$ & 86.97 & 52.23 & 87.04 & 65.28 \\
  & $\pm2$ & 84.82 & 51.76 & 91.09 & 66.02 \\
  \midrule
  \multirow{3}{*}{Local YETI}
  & 0 & 94.40 & 68.19 & 28.17 & 39.87 \\
  & $\pm1$ & 93.76 & 64.51 & 59.73 & 62.02 \\
  & $\pm2$ & 91.82 & 61.39 & 76.09 & 67.95 \\
  \bottomrule
  \end{tabular}
  \caption{Comparison over different Extrema ranges.}
  \label{tab:supp-extrema-interactions}
\end{table}

\subsection{SSIM Thresholding}

When we filter out certain frames based on their SSIM score relative to the previous frame, we are abstracting the meaning of what it means to have structural similarity. This abstraction is a form of reasoning.

\begin{table*}[!ht]
    \centering
    \setlength{\tabcolsep}{4pt}
    \begin{tabular}{c|c|ccc|ccc|ccc|ccc}
    \toprule
    \multirow{2}{*}{Method} & \multirow{2}{*}{SSIM} & \multicolumn{3}{c}{Overall} & \multicolumn{3}{|c}{Confirm Action} & \multicolumn{3}{|c}{Correct Mistake} & \multicolumn{3}{|c}{Follow Up} \\
     & & Prec. & Rec. & F-meas. & Prec. & Rec. & F-meas. & Prec. & Rec. & F-meas. & Prec. & Rec. & F-meas. \\
    \midrule
    \multirow{5}{*}{Global YETI}
     & 0.5 & 43.20 & 71.83 & 53.95 & 23.86 & 77.14 & 36.45 & 11.40 & 71.87 & 19.68 & 30.01 & 76.74 & 43.14\\
     & 0.6 & 42.55& 79.60 & 55.46 & 23.07 & 83.90 & 36.19 & 11.70 & 81.33 & 20.46 & 29.53 & 83.02 & 43.56\\
     & 0.7 & 41.71 & 85.49 & 56.06 & 22.56 & 88.93 & 35.99 & 11.88 & 87.77 & 20.93 & 28.81 & 87.69 & 43.36\\
     & 0.8 & 41.29 & 87.38 & 56.08 & 22.44 & 90.64 & 35.98 & 11.87 & 89.37 & 20.96 & 27.99 & 89.05 & 40.59\\
     & 0.9 & 41.86 & 88.31 & 56.17 & 22.54 & 91.51 & 36.17 & 11.68 & 90.29 & 20.69 & 27.55 & 89.69 & 42.15\\
     \midrule
     \multirow{5}{*}{Local YETI}
     & 0.5 & 47.35 & 43.05 & 45.09 & 25.88 & 50.31 & 34.18 & 14.23 & 45.22 & 21.65 & 32.66 & 48.00 & 38.87\\
     & 0.6 & 45.55 & 49.19 & 47.30 & 24.60 & 56.28 & 34.24 & 14.00 & 54.62 & 22.29 & 30.84 & 53.22 & 39.05\\
     & 0.7 & 45.93 & 55.86 & 50.41 & 25.09 & 63.09 & 35.91 & 14.50 & 63.14 & 23.59 & 30.44 & 58.83 & 40.12\\
     & 0.8 & 47.09 & 59.52 & 52.58 & 26.5 & 67.61 & 38.08 & 14.98 & 67.39 & 24.51 & 30.45 & 61.47 & 40.73\\
     & 0.9 & 46.88 & 60.38 & 52.77 & 26.55 & 68.62 & 38.29 & 14.71 & 68.02 & 24.18 & 30.07 & 62.05 & 40.51\\
    \bottomrule
    \end{tabular}
    \caption{Evaluation of YETI for SSIM Thresholding for proactive interventions}
    \label{tab:supp-ssim-interventions}
\end{table*}

\begin{table}[!ht]
  \centering
  \setlength{\tabcolsep}{4pt}
  \begin{tabular}{c|c|cccc}
  \toprule
  \multirow{2}{*}{Method} & \multirow{2}{*}{SSIM} & \multicolumn{4}{c}{Overall} \\

  &   & Acc. & Prec. & Rec. & F1 \\
  \midrule
  \multirow{5}{*}{Global YETI}
  & 0.5 & 91.49 & 55.38 & 68.88 & 61.39 \\
  & 0.6 & 90.20 & 54.14 & 77.36 & 63.69 \\
  & 0.7 & 88.62 & 53.06 & 83.92 & 65.01 \\
  & 0.8 & 87.55 & 52.39 & 85.99 & 65.11 \\
  & 0.9 & 86.97 & 52.23 & 87.04 & 65.28 \\
  \midrule
  \multirow{5}{*}{Local YETI}
  & 0.5 & 94.13 & 64.25 & 41.12 & 50.14 \\
  & 0.6 & 93.98 & 62.75 & 48.05 & 54.42 \\
  & 0.7 & 93.88 & 63.46 & 55.14 & 59.01 \\
  & 0.8 & 93.81 & 64.42 & 58.65 & 61.40 \\
  & 0.9 & 93.76 & 64.51 & 59.73 & 62.02 \\
  \bottomrule
  \end{tabular}
  \caption{Comparison over different SSIM Thresholding levels for proactive interactions.}
  \label{tab:supp-ssim-interactions}
\end{table}

\subsection{Episode Length}

Varying the initial history (i.e. Episode Length) needed in order to make the first proactive intervention/interaction did not influence the results from those in Tables \ref{tab:intervention-results} and \ref{tab:interaction-results}. This could be due to the fact that the Episode Length is only relevant for the first few frames of the video, and thus has an extremely narrow window to influence the ultimate output.

\section{Additional Use-Cases for Proactive Intervention Detection}

The HoloAssist paper presents many additional use cases for an AI Assistant that can proactively intervene when the user is trying to accomplish a task. An exhaustive list of all the uses included in the dataset is as follows:

\begin{itemize}
\item Assembly tasks
\begin{itemize}
    \item Assembling a computer
    \item Assembling a laser scanner
    \item Assembling a nightstand
    \item Assembling a stool
    \item Assembling a tray table
    \item Assembling a utility cart
    \item Disassembling a nightstand
    \item Disassembling a tray table
    \item Disassembling a utility cart
\end{itemize}
\item Maintenance and Repair
\begin{itemize}
    \item Changing a mechanical belt
    \item Changing a circuit breaker
    \item Fixing a motorcycle
\end{itemize}
\item Consumer Electronics
\begin{itemize}
    \item Making coffee with an espresso machine
    \item Making coffee with an espresso machine
    \item Setting up a big printer
    \item Setting up a camera
    \item Setting up a GoPro
    \item Setting up a small printer 
    \item Setting up a Nintendo Switch
\end{itemize}
\end{itemize}

Assistive AI Agents that can proactively intervene are particularly useful for tasks such as these, where they can help with:

\begin{enumerate}
    \item \textbf{Complexity Management:} Furniture assembly is a good example of a task that seems straightforward at first but can quickly get out of hand, especially if a mistake made early on is only noticed by the user much later on in the process. A proactive AI Agent intervention could help catch these mistakes early on before they become a larger problem.
    
    \item \textbf{Safety Considerations:} Tasks such as changing a mechanical belt or a circuit breaker can open the user up to serious harm if they are not careful. If the user interacts with a live wire while the power is on they could be seriously injured. An AI assistant could potentially proactively intervene before the user makes this mistake.
    
    \item \textbf{Expertise Gap:} If a user is new to completing a task that benefits from learned experience, the proactive AI Agent could take on the role of an instructor guiding a student. This would be especially useful for fixing a motorcycle, which is often done by expert mechanics.
\end{enumerate}

The diversity of these use-cases demonstrates that proactive AI assistance has broad applicability in scenarios where expert oversight would traditionally be beneficial. This is particularly valuable in contexts where users might not possess sufficient domain knowledge to self-identify potential issues or formulate appropriate queries for assistance.

\begin{figure*}[htbp]
    \centering
    \subfloat[Intervention - 3 seconds]{
        \includegraphics[width=0.23\textwidth]{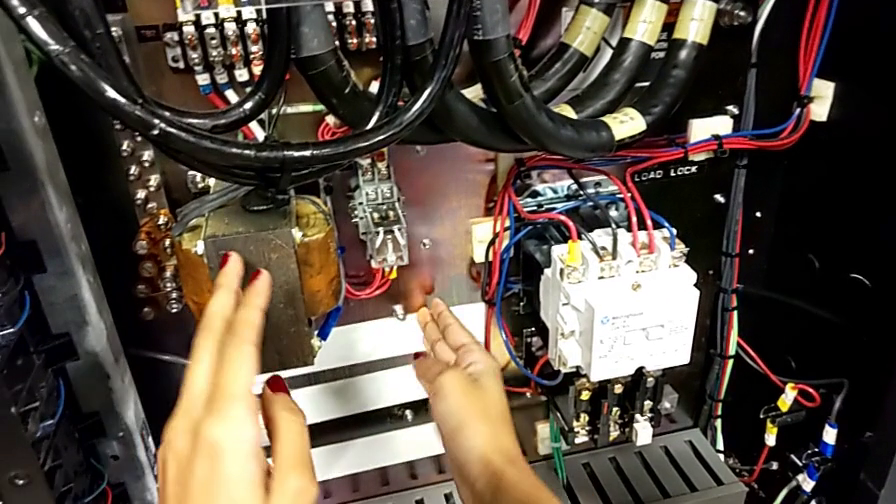}
        \label{fig:1}
    }
    \hfill
    \subfloat[Intervention - 2 seconds]{
        \includegraphics[width=0.23\textwidth]{Figures/change-circuit-breaker/frame40.png}
        \label{fig:2}
    }
    \hfill
    \subfloat[Intervention - 1 second]{
        \includegraphics[width=0.23\textwidth]{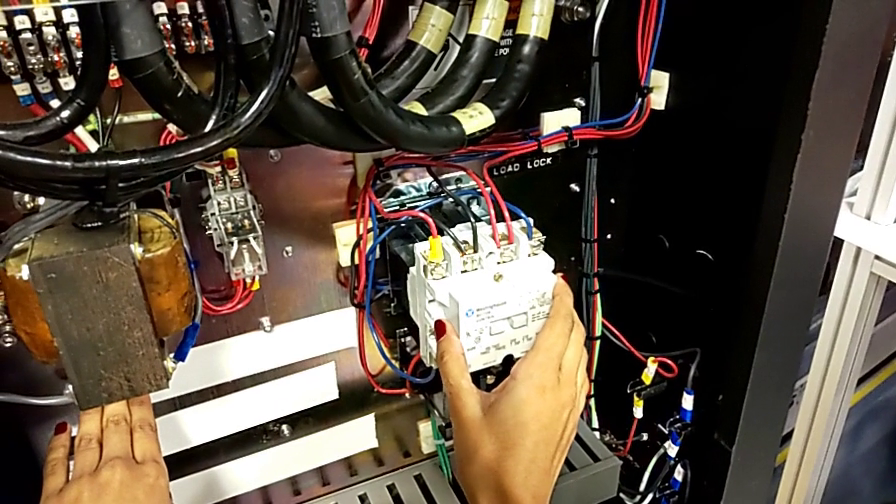}
        \label{fig:3}
    }
    \hfill
    \colorbox{green!30}{
    \subfloat[AI Agent proactively intervenes.]{
        \includegraphics[width=0.23\textwidth]{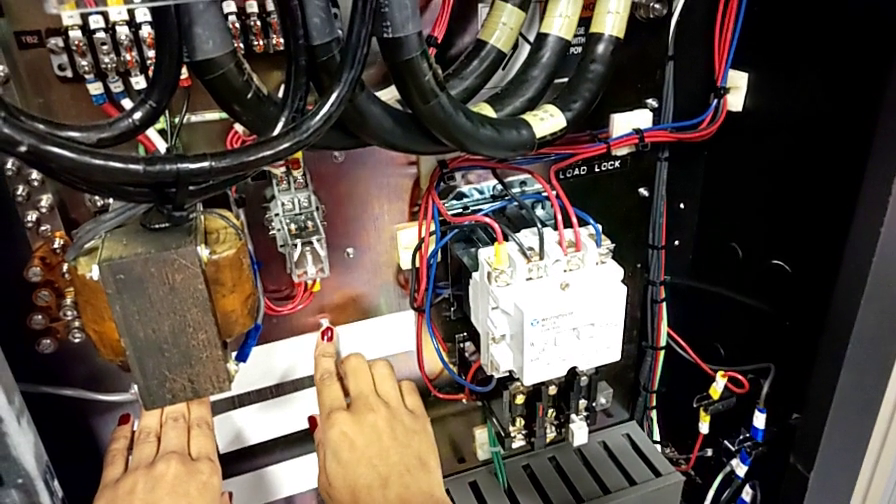}
        \label{fig:4}
    }
    }
    
    \vspace{1em}  
    
    \subfloat[SSIM between consecutive frames.]{
            \includegraphics[width=0.45\textwidth]{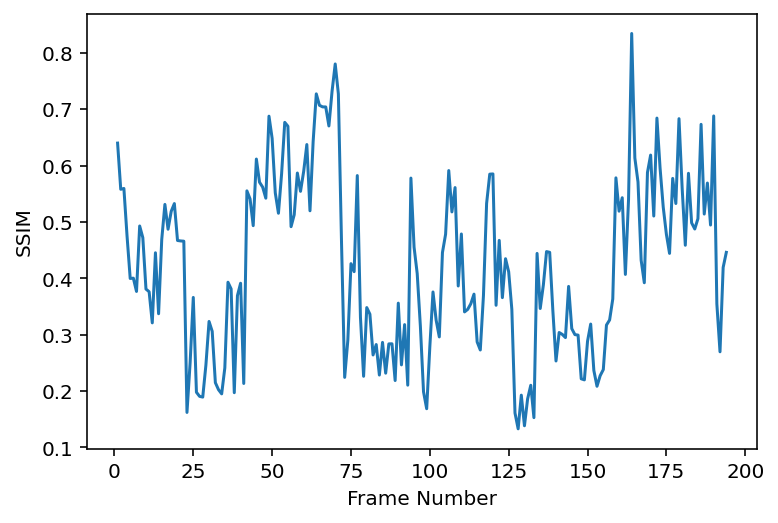}
        \label{fig:5}
    }
    \hfill
    \subfloat[Changing Objects Alignment Signal]{
        \includegraphics[width=0.45\textwidth]{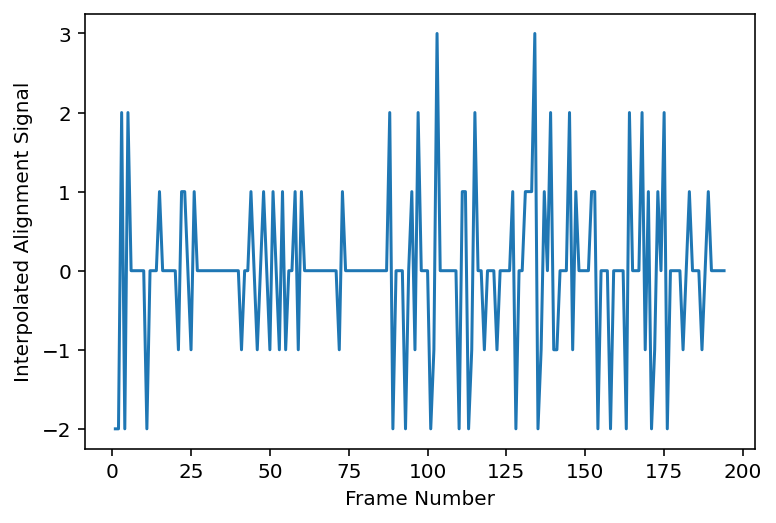} 
        \label{fig:6}
    }
    
    \caption{Intervention Detection for Changing Electric Circuit Task.}
    \label{fig:all-circuit}
\end{figure*}

A Proactive AI Agent can help in many industrial or home maintenance tasks like how to change an electrical task in Figure \ref{fig:all-circuit}. The AI Agent helps to guide the user on how to change the circuit breaker and proactively intervenes without any questions from the user when the AI Agent observes that the user may touch the electric circuit which will be a safety worry. The Extrema of the changing object count for the corresponding frame is helpful to determine when the AI Agent should intervene proactively, post filtering by the SSIM signal. Filtering helps to obtain an abstract scene understanding capability to the Proactive AI Agent. The expectation is that the frames where the AI Agent should intervene are dissimilar and can be observed by SSIM while pruning the edge cases by filtering. 
